\documentclass[conference]{IEEEtran}
\IEEEoverridecommandlockouts
\usepackage{cite}
\usepackage{amsmath,amssymb,amsfonts}
\usepackage{graphicx}
\usepackage{textcomp}
\usepackage{xcolor}
\usepackage{subcaption}
\usepackage{multicol}
\usepackage{booktabs}
\usepackage{comment}
\usepackage{multirow}
\usepackage{algpseudocode}
\usepackage{algorithm}
\usepackage{url}
\newcommand{\RomanNumeralCaps}[1]
    {\MakeUppercase{\romannumeral #1}}
\def\BibTeX{{\rm B\kern-.05em{\sc i\kern-.025em b}\kern-.08em
    T\kern-.1667em\lower.7ex\hbox{E}\kern-.125emX}}
\begin{document}

\title{Temporal Clustering with External Memory Network for Disease Progression Modeling

}

\DeclareRobustCommand*{\IEEEauthorrefmark}[1]{%
  \raisebox{0pt}[0pt][0pt]{\textsuperscript{\footnotesize\ensuremath{#1}}}}

\author{
    \IEEEauthorblockN{Zicong Zhang\IEEEauthorrefmark{1}, Changchang Yin\IEEEauthorrefmark{1}\textsuperscript{,}\IEEEauthorrefmark{2}, Ping Zhang\IEEEauthorrefmark{1}\textsuperscript{,}\IEEEauthorrefmark{2}}
    \IEEEauthorblockA{ \IEEEauthorrefmark{1}Department of Computer Science and Engineering, The Ohio State University, Columbus, USA \\
                         \IEEEauthorrefmark{2}Department of Biomedical Informatics, The Ohio State University, Columbus, USA \\
    Email: \{zhang.5157, yin.731, zhang.10631\}@osu.edu}
}
\maketitle

\begin{abstract}

Disease progression modeling (DPM) involves using mathematical frameworks to quantitatively measure the severity of how certain disease progresses. DPM is useful in many ways such as predicting health state, categorizing disease stages, and assessing patients' disease trajectory, etc. Recently, with the wider availability of electronic health records (EHR) and the broad application of data-driven machine learning methods, DPM has attracted much attention yet remains two major challenges: (i) Due to the existence of irregularity, heterogeneity, and long-term dependency in EHRs, most existing DPM methods might not be able to provide comprehensive patient representations. (ii) Lots of records in EHRs might be irrelevant to the target disease. Most existing models learn to automatically focus on the relevant information instead of explicitly capture the target-relevant events, which might make the learned model suboptimal. To address these two issues, we propose Temporal Clustering with External Memory Network (TC-EMNet) for DPM that groups patients with similar trajectories to form disease clusters/stages. TC-EMNet uses a variational autoencoder (VAE) to capture internal complexity from the input data and utilizes an external memory work to capture long-term distance information, both of which are helpful for producing comprehensive patient health states. Last but not least, the k-means algorithm is adopted to cluster the extracted comprehensive patient representation to capture disease progression. Experiments on two real-world datasets show that our model demonstrates competitive clustering performance against state-of-the-art methods and is able to identify clinically meaningful clusters. The visualization of the patient representations shows that the proposed model can generate better patient health states than the baselines. 

\end{abstract}

\begin{IEEEkeywords}
disease progression modeling, deep learning, temporal clustering
\end{IEEEkeywords}

\section{Introduction}

With the recent development of deep learning and the accumulation of electronic health records (EHR), also known as time-series data, there has been an increasing effort in clustering EHR data in order to discover meaningful patterns throughout longitudinal health information. Moreover, chronic diseases, such as Parkinson's disease (PD) and Alzheimer's disease (AD), can have various outcomes even with a limited number of patients. Such diseases are heterogeneous in nature and often evolve at unique patterns that trigger distinct responses to therapeutic interventions based upon different conditions \cite{kehagia2010neuropsychological}. Thus, it has become crucial to develop a disease progression modeling (DPM) system to capture certain progression patterns, provide early detection to critical situations, and yield clinically helpful information to improve the quality of care.


Traditionally, DPM or disease clustering/staging is developed by domain experts with extensive clinical experience, in which disease stages are defined separately and based solely on the values of one or a few biomarkers \cite{ferrer1997chronic,auer1997gds}. Nevertheless, developing a DPM system requires long-term observation and human labor, and the result is often based on known biomarkers and acknowledged covariants, which makes it difficult to develop a DPM system for disease with limited knowledge on biomarkers that have not been well-studied. In recent years, the rapid growth of data-driven machine learning methods has motivated a great effort in developing DPM models. There are two main approaches when it comes to DPM: 1) The problem is formed as a risk prediction task with label information based on patient representation that is extracted from the last layer of the model.  \cite{lee2020temporal,gao2020stagenet,ma2018health,sun2019probabilistic,zhang2019data}. 2) The problem is formed as a traditional unsupervised, patient clustering/subtyping problem where the model is trained to separate the patient into multiple groups \cite{wang2014unsupervised,fortuin2018som,mou2019t}. Leveraging disease outcomes during the training process can prevent the model from forming heterogeneous clusters. However, for certain diseases, diagnosis labels are often unavailable at each patient visit due to limited knowledge of the disease. Moreover, deep learning models that are designed for supervised tasks may not perform well when training in an unsupervised fashion. Therefore, there is a need for developing a DPM framework that can handle both situations with respect to the availability of training labels. However, most developed deep learning models for disease progression modeling suffers from the following limitations:
\begin{itemize}
  \item \textbf{Irregularity and heterogeneity}: Many diseases are heterogeneous in nature and EHR data often has high internal complexity. Due to the complexity of effectively encoding various health conditions into patient representation, accurate DPM still remains a challenging problem.
  \item \textbf{Long-term Dependency}: RNNs are long known to suffer from modeling long-term dependency since it tends to forget earlier information when the input sequence is long. Disease progression modeling, especially for chronic disease, requires long-term observation of the patient in order to provide a comprehensive view for decision making.
  \item \textbf{Target Awareness}: Most rnn-based methods derive patient representations directly from the hidden states of the model. Such an approach neglect the contribution of target-relevant information. In fact, real-world clinical decisions made by doctors are often based upon past diagnoses as well. 
\end{itemize}

To address these challenges, we propose Temporal Clustering with External Memory Network (TC-EMNet) for disease progression modeling via both supervised and unsupervised settings. TC-EMNet leverages a variation autoencoder framework and a memory network to deal with data irregularity and long-term dependency problems of RNNs respectively. At each time step, TC-EMNet takes EHR medical records as input and encodes the input feature using a recurrent neural network to get hidden representations. Then TC-EMNet samples from the hidden state to form a latent representation. Meanwhile, the hidden state is stored in a global-level memory network, which in turn outputs a memory representation based on current memory cells. The memory representation is then concatenated with the current latent representation to form the patient representation at the current time step. When the training label is available, the model also employs a patient-level memory network to process label information up to most recent visit and outputs target-aware memory representation. We combine memory representations from global-level and patient-level memory networks using a calibration process. TC-EMNet is trained with reconstruction objective under unsupervised setting and prediction objective under supervised setting.

In this paper, our contributions are four fold:
\begin{itemize}
  \item We propose a novel deep learning framework, namely TC-EMNet for disease progression modeling under both supervised and unsupervised settings.
  \item TC-EMNet uses a combined recurrent neural network and variational auto-encoder (VAE) architecture to capture the irregularity in data and heterogeneity nature of the disease.
  \item Under superviesd setting, TC-EMNet employs dual memory network architecture to leverage both hidden representations from the input data and clinical diagnosis to produce accurate patient representations. 
  \item Experiments on two world datasets show that TC-EMNet yields competitive clustering performance over state-of-the-art methods and is able to find clinically interpretable disease clusters/stages.
\end{itemize}

The remainder of the paper is organized as follows. Section II briefly reviews existing works related to DPM, temporal clustering, and VAE. Section III describes the technical details of the proposed model (TC-EMNet). Section IV and V present experimental results and discussions. Finally, Section VI concludes the paper.

\begin{figure*}[t]
  \centering
  \includegraphics[width=0.85\textwidth]{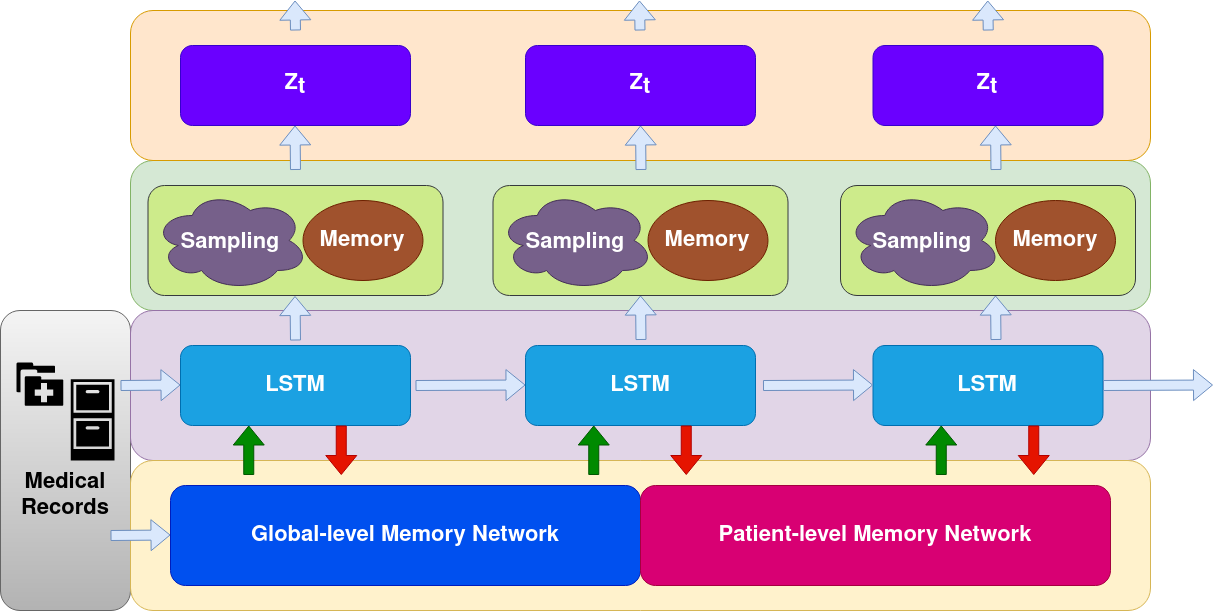}
  \caption{
  Overview of the proposed framework. At each timestamp, hidden representation from the encoder network is updated with the memory state to produce disease clusters/stages based on current and previous observations.
  }
  \label{framework}
\end{figure*}

\section{Related Work}

\subsection{Disease Progression Modeling}

Disease progression modeling (DPM) plays a very important role in the healthcare domain, especially for chronic diseases such as Parkinson's Disease (PD) and Alzheimer's Disease (AD). A well-performed disease progression modeling system can not only provide early detection or diagnosis but also discover clinically meaningful patterns for certain groups of trajectories.
Most probabilistic models for DPM are based on the hidden markov model (HMM). 
For example, \cite{alaa2019attentive} derived a deep probabilistic model based on sequence-to-sequence architecture to model progression dynamic on UK Cystic Fibrosis Registry. \cite{wang2014unsupervised} introduced a continuous-time Markov process to learn a discrete representation of each progression state. Moreover, deep learning methods have also been developed for disease progression modeling. \cite{liu2018joint} proposed a CNN-based model to jointly learn features from MR images combined with demographic information to predict Alzheimer's Disease progression patterns. \cite{teng2020stocast} designed a prediction framework using generative models to forecast the distribution of patients' outcomes. DPM can be regarded as a classification problem, where diagnosis labels are leveraged in favor of model training. On the other hand, DPM can also be seen from an unsupervised perspective where the goal is to discover potential disease states or patient subtypes throughout patients' medical history \cite{dennis2019disease}. However, DPM still remains a challenge due to the high complexity of data introduced by irregular progression patterns for certain chronic diseases.

\subsection{Temporal clustering}

Temporal clustering, widely known as time-series clustering, is a data-driven method to cluster patients into subgroups based on time-series observation. Temporal clustering can be considered a challenging task often because of the high dimensionality of the dataset and multiple time steps for each data sample. Recent advances have been focused on leveraging the latent representation learned by recurrent neural network (RNN) for temporal clustering, which was motivated by the success of RNN modeling time-series data. Moreover, due to the emerging availability of electronic health records (EHR) that introduced large-scale and normalized context for individual patients, the deep learning approach become capable of learning more comprehensive patterns and achieving better performance on several critical tasks. \cite{baytas2017patient} introduces a time-aware mechanism to long short term memory cells to capture progression patterns with irregular time-interval. \cite{lee2020temporal} proposed an actor-critic algorithm for predictive clustering where, instead of defining a similarity measure for clustering, a cluster embedding is trained to represent each disease stage. \cite{yin2020identifying} proposed an auto-encoder to reconstruct relevant features for sepsis with attention and showed that the proposed model can identify interpretable patient subtypes. Nevertheless, there is only limited literature that focuses on DPM using temporal clustering techniques.

\subsection{Variational Autoencoder}

Variational autoencoder (VAE) is a type of generative model that can handle complicate distributions. VAEs are effective against modeling complex data structures and are widely adopted to solve many real-world problems range from image generation to anomaly detection \cite{burgess2018understanding, higgins2016beta, kuo2017variational}. It has also several successful applications with healthcare data \cite{shickel2017deep}. \cite{jun2019stochastic} proposed to use VAE to impute missing values for electronic health data with uncertainty-aware attention. Experiments on real-world datasets show that VAE can capture the complexity of EHR distribution. \cite{teng2020stocast} leveraged the VAE framework to forecast disease states for Parkinson's Disease (PD) and Alzheimer's Disease (AD). Nonetheless, the latent representation learned from VAE can be drawn from unrealistic distribution if trained without any constraints.

\section{Methodology}

\subsection{Problem Definition}

Let $x \subseteq \boldsymbol{X} $ and $y \subseteq \boldsymbol{Y} $ be the random variables for input feature space and label space accordingly. Here we focus on a clustering problem, where we are given a population of time-series data
$D = \{(x_{t}^{n}, y_{t}^{n})_{t=1}^{T}\}_{n=1}^{N}$ consisted of paired sequences of observations $(x, y)$ for $N$ patients. $t \subseteq {1, ..., T}$ denotes the time stamps for each patients at which the observations are made. 

We aim to identify $K$ clusters for time-series data, each corresponding to a disease stage. Each cluster consists of homogeneous data samples, represented by the centroids based on certain similarity measures.

\subsection{Method}

This section presents our proposed framework. Here we discuss disease progression modeling under both supervised and unsupervised settings, where our proposed question requires estimating the underlying distribution of all possible disease stages. Such a DPM framework can help the doctors identify meaningful characteristics in both times when a disease has certain diagnosis labels but possible underlying disease stages and when a disease has no well-defined diagnosis labels. 

The framework consists of three components: the encoder, the memory network, and the clustering network. For each patient, a recurrent neural network is deployed to encode the patient's information. The memory network controls the overall long-term information at each timestamp. Specifically, when a hidden representation $h_{i}$ is generated based on current and previous observation $X_{<i}$ at timestamp $i$, the hidden state is read by the memory network and updates the memory storage. Next, a latent variable $z_{i}$ is drawn from the prior distribution $p_{\theta}(z_{i}|X_{<i})$ conditioned on the hidden state that is generated from the memory network. Then, we either yield prediction outcomes or reconstruct the current observation $X_{i}$ accordingly. We take the hidden presentation from the last layer of the model for clustering.

\subsubsection{Encoder Network}

The encoder network takes the current observation and the hidden state from the previous timestamp and yields the hidden representation that can interact with the external network. Specifically, a LSTM cell is adopted to generate and update the hidden state:

\begin{equation}
    h_{t} = LSTM(X_{t}, h_{t-1}),
    \label{encoding}
\end{equation}

where $X_{t}$ is the current observation at timestamp $t$ and $h_{t-1}$ is the hidden state from previous step. At each timestamp, the encoder network maps a sequence of time-series input $x_{1:t}$ to a hidden representation $z_{t} \subseteq Z$, where $Z$ is the subspace of latent representation. The hidden representation will be interacting with the external memory network to form an accurate representation.

\subsubsection{Memory Network}

Long-term information plays an important role in disease progression modeling, since, in the context of chronic disease, the health conditions from the past will affect the current disease stages of the patient. In addition, historical information should be stored in an efficient way such that it can provide useful guidance towards the patient's current health state at different timestamps. To this end, we propose an external memory network to capture long-term information throughout the progression modeling process. Our proposed memory network is closely related to \cite{sukhbaatar2015end}, which has several successful applications in the field of natural language processing. Similarly, we define memory slots to represent historical information that can be extracted and retrieved at any given timestamp. At each timestamp, the hidden state from the encoder network is recorded and written into the memory cells. By pushing through a series of observations, the memory network will process continuous representations for each individual visit so that a more comprehensive review of the patient can be utilized during the clustering/staging process.

\paragraph{Memory Reading}

We denote a clinical sequences record $r_{t}, t=1,...,T$, where t stands for index or timestamps of the given record. In memory network, after receiving a hidden representation $z_{t}$ from the encoder network, the network will produce an external representation $e_{t}$ based on reading weight $w_{t}^{l}, l=1,...,T$ of the memory slots. Specifically, $e_{t}$ can be expressed as:
\begin{equation}
    \begin{aligned}
      &  e_{t} = \sum_{l} w_{t}^{l}m_{t}, \\
      &   w_{t}^l = \textbf{softmax}(\alpha_{t}, m_{t}, h_{t}) \\
      &  = \frac{\exp{(\alpha_{t}^{l})}C(h_t, \mathbf{M}_{t})}{\sum_{j} \exp{(\alpha_{t}^{j})}C(h_t, \mathbf{M}_{t})}
    \end{aligned}
     \label{memory_read}
\end{equation}

where $l=1,...L$ denotes the number of memory slots, $m_{t}\subseteq\mathbb{R}^{1 \times D}$ is the memory representation with hidden size $D$. $\alpha$ is the strength vector that can be learned through the reading operation and $C(\dots)$ is the cosine similarity measure. Memory reading operation is built upon the idea that not all records in the sequence contribute equally to the current health state of the patient. Hence, the weights are computed using the softmax function based on the cosine similarity of the current hidden states and all the previous memories. 

\begin{figure}[t]
  \centering
  \includegraphics[width=\linewidth]{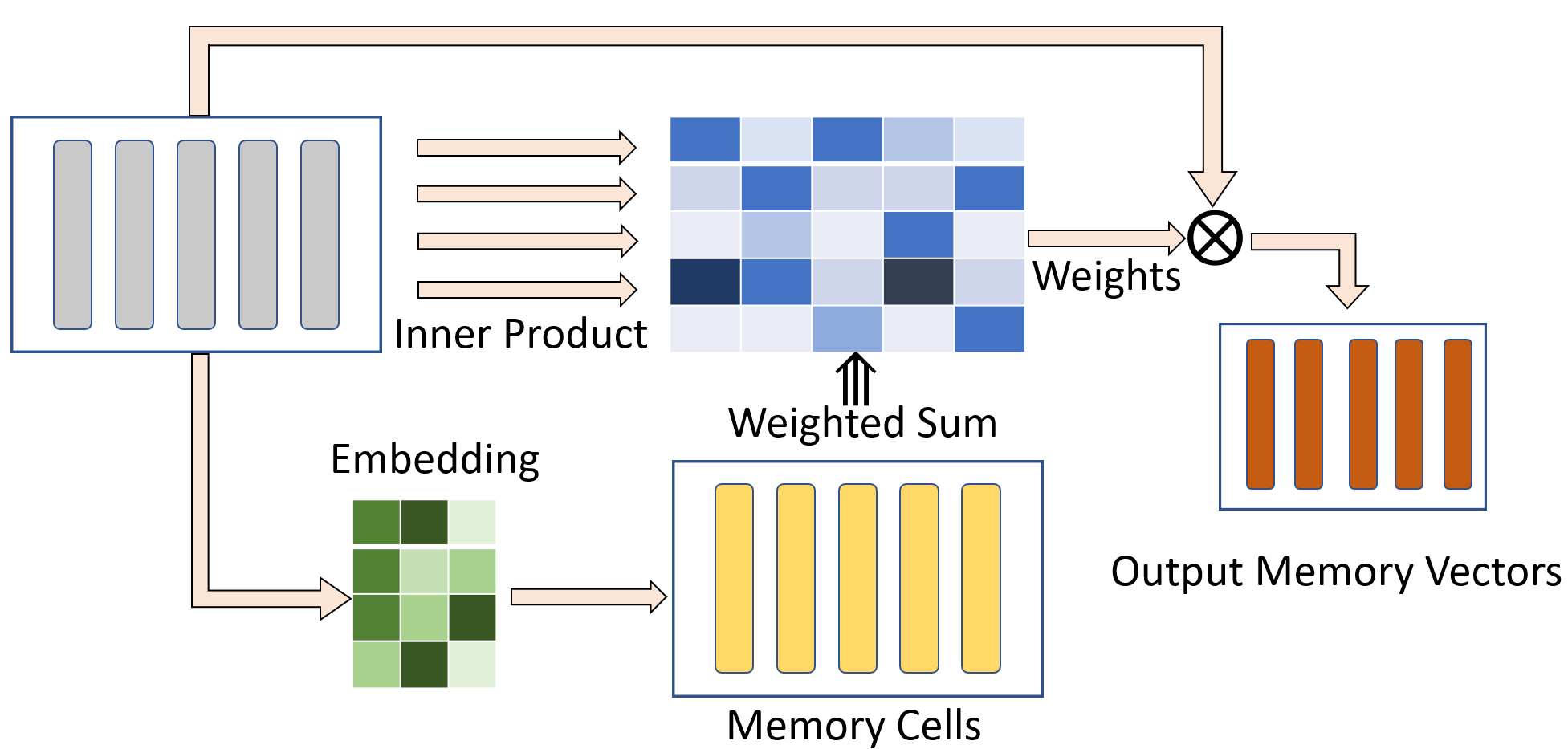}
  \caption{
  Overview of the proposed memory network. Hidden states are first written into the memory cells and read by the clustering network to produce a comprehensive representation.
  }
  \label{memory}
\end{figure}

\paragraph{Memory Writing}

Memory writing stores latent representation into memory slots. We use a fixed number of slots to denote the overall memory size. The dimension of the continuous space for each memory slot is $d$ and we use $D$ to denote the dimension of hidden representation $z_{t}$. The hidden state is non-linearly projected into the memory space using a $d\times D$ matrix \textbf{A}, $m_{t} = Az_{t}$, where $m_{t}$ is the new input memory representation. Memory writing aims to filter out non-related information and stores only personalized information based on the current hidden state. Mathematically, memory writing can be expressed as:
\begin{equation}
    \begin{aligned}
        \mathbf{M}_{t} = r_{t}\mathbf{M}_{t-1} +  v_{t}h_{t},
    \end{aligned}
    \label{memory_write}
\end{equation}

where $r_{t}$ and $v_{t}$ are gated vectors that control the information flow between the previous and current memory vector.

\subsubsection{Clustering Network}

After obtaining the representation of the observation $h_{t}$ through the encoding network, i.e the prior network, and updating the memory cell $m_{t}$ at current timestamp $t$, we follow the traditional framework of variational autoencoder (VAE) \cite{lopez2017conditional} to compute the mean and standard deviation vectors through the posterior network. We assume that the output is a Gaussian distribution. The computation process can be expressed as:
\begin{equation}
    \begin{aligned}
      & \mathbf{\mu}_{z_{t}} = f_{posterior}([h_{t}, x_{i}]) \\
      & \mathbf{\sigma}_{z_{t}} = f_{posterior([h_{t}, x_{i}])}
    \end{aligned}
    \label{vae}
\end{equation}
where $h_{t}$ is the hidden state and $x_{i}$ is the observation at timestamp $t$. $f_{posterior}$ is posterior functions described by feed-forward neural networks. We then draw samples from the posterior Gaussian distribution using the reparameterization trick:
\begin{equation}
    z_{t} = \mu_{z_{t}} + \sigma_{z_{t}}\odot\epsilon,
\end{equation}
where $\epsilon \sim \mathcal{N}(0, 1)$, and $z_{t}$ is the latent representation. $\odot$ indicates element-wise multiplication. The reparameterization trick allows the gradient to backpropagate through the sampling process. Lastly, depends on the availability of diagnosis labels, the clustering network will be trained on two different objectives. When diagnosis label is used, the clustering network is directly trained to predict the label information:
\begin{equation}
    \hat{Y_{t}} = f_{pred}(f_{con}([z_{t}, m_{t}])),
\end{equation}
where $f_{pred}$ is a feed-forward network that outputs probabilities of each label. When diagnosis label is not available, we trained the framework to reconstruct the observation $x_{i}$ from the latent variable $z_{t}$ conditioned on the memory state $m_{t}$, denote as:
\begin{equation}
    \hat{X_{t}} = f_{recon}(f_{con}([z_{t}, m_{t}])),
    \label{recon}
\end{equation}
where $\hat{X_{t}}$ is the reconstructed input, $f{recon}$ is a feed forward network and $f_{con}$ is the concatenation. During cluster phase, we use euclidean distance-based k-means algorithm on the latent variable $z_{t}$.
\begin{algorithm}[t]
    \caption{TC-EMNet}
    \label{algo}
    \begin{algorithmic}[1]
    \item Initialize encoder and decoder network parameters $\theta, \sigma$;
    \item Initialize memory embedding and memory slots;
     \For {(every time stamp $t$)}
         \State Compute patient hidden encoding through Encoder 
         \State network via  Eq. (\ref{encoding});
         \State Read from global-level memory network to extract 
         \State recent memory representations via Eq. (\ref{memory_read});
        \If{diagnosis is available}    
        \State Read from patient-level memory network 
        \State to extract recent memory representations 
        \State via Eq. (\ref{memory_read});
        \State Compute memory representation via Eq. (\ref{cali});
        \EndIf
        \State Compute loss via Eq. (\ref{vae}) - (\ref{recon});
        \State Write to corresponding memory slots via Eq. (\ref{memory_write});
     \EndFor   
     \item Update parameters by optimizing Eq. (\ref{objective1}), (\ref{objective2})\ accordingly;
    \end{algorithmic}
\end{algorithm}
\subsubsection{Dual Memory Network Architecture}

Under a clinical setting, doctors often provide diagnosis labels based on patients' current and past medical events. Such information can be target health conditions or a diagnosis. Under supervised setting when the label is available during training, we further utilize a patient-level memory network to capture diagnosis information during each visit. Compared to global-level memory network, patient-level memory network at current memory slot $M_{t}$ can only access diagnosis up to previous timestamp, namely,  $\{Y_{i} | i = 1,2,...,t-1\}$. patient-level memory network only reads and writes diagnosis information which later is combined with a global-memory network for clustering. We propose a calibration process to integrate representations from two memory networks, as follows:
\begin{equation}
    \begin{aligned}
        & h_{t}^{global} = M_{global}(X; \alpha_{t}^{global}, m_{t}^{global}, h_{t}^{global}), \\
        & h_{t}^{patient} = M_{patient}(Y_{<t};  \alpha_{t}^{patient}, m_{t}^{patient}, h_{t}^{patient}), \\
        & h_{t}^{final} = h_{t}^{global} \odot \sigma(f_{embed}(h_{t}^{patient} )), 
    \end{aligned}
    \label{cali}
\end{equation}
where $M_{global}$ and $M_{patient}$ is the global-level and patient-level memory network respectively. This memory calibration process can be regarded as a point-wise attention mechanism. 
\subsection{Objective Function and Optimization}

Here, we present our training objectives and optimization process. As mentioned in previous sections, the entire network can be trained from end to end using maximum likelihood estimation (MLE). To solve the intractable marginalization for the latent variable $z_{i}$, we use the variational lower bound parameterized by $q_{\phi}$ to approximate the true distribution, which we assume to be Gaussian. After the memory work reading and writing, We use the latent variable $z_{i}$ at timestamp $i$ to identify the disease stages. Here we restrict the latent variable to be a multivariate Gaussian distribution, which enforces the same for the posterior.  We learn the generative parameter $\theta$ using maximum likelihood estimation (MLE):

\begin{equation}
    \begin{aligned}
        \theta = \operatorname*{argmin}_\theta \sum_{i} \log \int [ p_{\theta}(x_{i} | z_{i}, X_{<i}) \\
        \times p_{\theta}(z_{i} | X_{i}) ] \,dz_{i}
    \end{aligned}
\end{equation}

However, the marginalization of $z_{i}$ is intractable for complicated functions (for instance neural networks). Thus, we need to derive a variational lower bound (i.e. variational Bayesian method) to approximate the logarithm of the marginal probability of the observation, which is as follows:

\begin{equation}
    \begin{aligned}
      &  \log p_{\theta} (x_{i} | z_{i}, X_{<i}) \\
      &  = \log (\mathbf{E}_{q\phi} [\frac{p_{\theta}(x_{i},  z_{i} | X_{<i})}{q_{\phi}(z_{i} | x_{i}, X_{<i})])} \\
      &  \geq \mathbf{E}_{q\phi} [\log p_{\theta}(x_{i}, z_{i} | X_{<i})] \\
      &  - \mathbf{E}_{q\phi} [\log q_{\phi}(z_{i} | x_{i}, X_{<i})],
    \end{aligned}
\end{equation}

where the inequality can be obtained using Jensen's inequality and the variational lower bound involves the probability $q_{\phi}$ that are parameterized by $\phi$, which eventually approximate the intractable true posterior distribution $ p_{\theta}(z_{i} | X_{i})$. Since health-related data is often associated with high-dimensional and general more complicated distribution, we introduce the latent variable $z_{t}$ to capture the internal stochasticity from the data. We can train the entire clustering network end-to-end using stochastic optimization techniques. After obtaining the variational lower bound, the optimization follows the KL divergence that is the difference of log-likelihood and the variational lower bound:
\begin{equation}
    \begin{aligned}
      & \mathcal{L}_{variation}(\theta, \phi) \\
      &= -KL[q_{\phi}(z_{t}|h_{t}, X_{<i}) || p_{\theta}(z_{t}|h_{t}, X_{<i})] \\
      & = \textbf{E}_{\log q_{\phi}(z_{t}|z_{t}, X_{<i})} 
       - \textbf{E}_{\log p_{\theta}(z_{i}, x_{i}|X_{<i})}
    \end{aligned}
    \label{objective3}
\end{equation}

where $\phi$ and $\theta$ represents the model parameter and proxy posterior accordingly. The equation holds if the distribution of $q_{\phi}$ is equal to the true distribution. When diagnosis label is used during training, we use the cross-entropy loss to directly predict the outcome from the combined latent representation denoted as:
\begin{equation}
    \begin{aligned}
      & \mathcal{L}_{objective}(\theta,\phi,x_{t}) = \\
      & \alpha\mathcal{L}_{variation}(\theta,\phi) + \mathcal{L}_{CE}(Y, \hat{Y}; \theta,\phi,x_{t}),
    \end{aligned}
    \label{objective1}
\end{equation}

When the model is trained in a unsupervised manner, the overall objective function combined with the reconstruction loss becomes:
\begin{equation}
    \begin{aligned}
      & \mathcal{L}_{objective}(\theta,\phi,x_{t}) = \\
      & \alpha\mathcal{L}_{variation}(\theta,\phi) + \mathcal{L}_{recon}(X, \hat{X};\theta,\phi,x_{t}),
    \end{aligned}
    \label{objective2}
\end{equation}
where we use the mean square error (MSE) for reconstruction loss and $\alpha$ is a hyperparameter to prevent VAE from KL vanishing problem. We adopt a linear annealing schedule for $\alpha$ based on training steps denoted as:
\begin{equation}
    \begin{aligned}
      \alpha = \min(1, \frac{\textit{training step}}{x}),
    \end{aligned}
    \label{objective3}
\end{equation}
where $x$ is a threshold value. Last but not least, we use the k-means algorithm \cite{wagstaff2001constrained} on the patient representation to perform clustering. 
\begin{table*}[!ht]
\centering
\caption{Results of proposed methods and other methods on ADNI datasets. $\downarrow$ indicates that the smaller the better (0=best, and 1=worst). $\uparrow$ indicates that the greater the better (0=worst, and 1=best).}
\label{ADNI_results}

\begin{tabular}{lcccccc}
\hline
 \multicolumn{1}{l}{} & \multicolumn{3}{c}{w/o label} & \multicolumn{3}{c}{with label} \\ 
 \hline
Model & Purity $\uparrow$ & NMI $\uparrow$ & RI $\uparrow$ & Purity $\uparrow$ & NMI $\uparrow$ & RI $\uparrow$\\
\hline
RNN & 0.6799$\pm$0.00 & 0.1415$\pm$0.01 & 0.1406$\pm$0.02 & 0.8532$\pm$0.00 & 0.4020$\pm$0.01 & 0.3805$\pm$0.01\\
Bi-LSTM & 0.6810$\pm$0.02 & 0.1540$\pm$0.02 & 0.1559$\pm$0.02 & 0.8674$\pm$0.00 & 0.4092$\pm$0.01 & 0.4042$\pm$0.02\\
RETAIN & 0.6903$\pm$0.02 & 0.1787$\pm$0.01 & 0.1671$\pm$0.01 & 0.7144$\pm$0.02 & 0.2572$\pm$0.01 & 0.1838$\pm$0.03\\
Dipole & 0.6839$\pm$0.00 & 0.1707$\pm$0.01 & 0.1452$\pm$0.00 & 0.8904$\pm$0.01 & 0.4674$\pm$0.01 & 0.4776$\pm$0.02\\
StageNet & 0.6943$\pm$0.01 & \bfseries{0.2002}$\pm$0.01 & 0.1791$\pm$0.01 & 0.8513$\pm$0.01 & 0.4045$\pm$0.03 & 0.3744$\pm$0.01\\
AC-TPC & - & - & - & 0.8214$\pm$0.03 & 0.3362$\pm$0.07 & 0.3827$\pm$0.09\\
\hline
VAE & 0.6651$\pm$0.02 & 0.1023$\pm$0.02 & 0.1117$\pm$0.02 & 0.6495$\pm$0.04 & 0.1718$\pm$0.05 & 0.1042$\pm$0.04\\
Memory Network & 0.6887$\pm$0.02 & 0.1392$\pm$0.01 & 0.1584$\pm$0.02 & 0.8262$\pm$0.01 & 0.3603$\pm$0.01 & 0.3538$\pm$0.02\\

$\text{TC-EMNet}^{-u}$ & \bfseries{0.7040}$\pm$0.01 & 0.1967$\pm$0.02 & \bfseries{0.1891}$\pm$0.02 & 0.8904$\pm$0.00 & 0.4679$\pm$0.01 & 0.4889$\pm$0.01\\
$\text{TC-EMNet}^{-s}$ & - & - & - & \bfseries{0.9126}$\pm$0.01 & \bfseries{0.4789}$\pm$0.01 & \bfseries{0.4923}$\pm$0.02\\
\hline
\end{tabular}
\end{table*}

\begin{table*}[!h]
\centering
\caption{Results of proposed methods and other methods on PPMI datasets. $\downarrow$ indicates that the smaller the better (0=best, and 1=worst). $\uparrow$ indicates that the greater the better (0=worst, and 1=best).}
\label{PPMI_results}

\begin{tabular}{lcccccc}
\hline
 \multicolumn{1}{l}{} & \multicolumn{3}{c}{w/o label} & \multicolumn{3}{c}{with label} \\ 
 \hline
Model & Purity $\uparrow$ & NMI $\uparrow$ & RI $\uparrow$ & Purity $\uparrow$ & NMI $\uparrow$ & RI $\uparrow$\\
\hline
RNN & 0.7221$\pm$0.00 & 0.3089$\pm$0.01 & 0.3120$\pm$0.01 & 0.7640$\pm$0.02 & 0.4222$\pm$0.04 & 0.3663$\pm$0.03\\
Bi-LSTM & 0.7264$\pm$0.00 & 0.3170$\pm$0.00 & 0.2976$\pm$0.01 & 0.7674$\pm$0.03 & 0.4456$\pm$0.05 & 0.3575$\pm$0.05\\
RETAIN & 0.5241$\pm$0.02 & 0.1188$\pm$0.01 & 0.0619$\pm$0.01 & 0.7510$\pm$0.01 & 0.4072$\pm$0.03 & 0.3361$\pm$0.01\\
Dipole & 0.7233$\pm$0.00 & 0.3200$\pm$0.00 & 0.3153$\pm$0.00 & 0.8033$\pm$0.01 & 0.4947$\pm$0.01 & 0.4476$\pm$0.02\\
StageNet & 0.7252$\pm$0.01 & 0.3305$\pm$0.00 & 0.3234$\pm$0.01 & 0.7839$\pm$0.01 & 0.4700$\pm$0.03 & 0.3840$\pm$0.01\\
AC-TPC & - & - & - & 0.8151$\pm$0.01 & 0.4984$\pm$0.03 & \bfseries{0.5129}$\pm$0.01\\
\hline
VAE & 0.7161$\pm$0.00 & 0.3576$\pm$0.01 & 0.3153$\pm$0.00 & 0.7942$\pm$0.01 & 0.4452$\pm$0.00 & 0.3782$\pm$0.01\\
Memory Network & 0.6996$\pm$0.01 & 0.2809$\pm$0.01 & 0.2581 $\pm$0.02 & 0.7689$\pm$0.01 & 0.4482$\pm$0.01 & 0.4597$\pm$0.01\\

$\text{TC-EMNet}^{-u}$ & \bfseries{0.7452}$\pm$0.00 & \bfseries{0.3773}$\pm$0.00 & \bfseries{0.3742}$\pm$0.01 & 0.8256$\pm$0.00 & \bfseries{0.5053}$\pm$0.00 & 0.4823$\pm$0.01\\
$\text{TC-EMNet}^{-s}$ & - & - & - & \bfseries{0.8339}$\pm$0.00 & 0.5035$\pm$0.00 & 0.4993$\pm$0.01\\
\hline
\end{tabular}
\end{table*}

\section{Experiments}

We evaluated our proposed model on two real-world datasets, Alzheimer's Disease Neuroimaging Initiative (ADNI) and Parkinson's Progression Markers Initiative (PPMI) dataset. All dataset can be accessed on IDA website\footnote[1]{\url{https://ida.loni.usc.edu/}}. The code can be found on GitHub\footnote[2]{\url{https://github.com/Ericzhang1/TC-EMNet.git}}. 

\subsection{Datasets}

\subsubsection{ADNI Dataset}

Alzheimer's disease (AD) is a chronic neurodegenerative disease that is often related to behavior and cognitive impairment. ADNI is a longitudinal study that aims to explore early detection and tracking of AD based on imaging, biomarkers, and genetic data collected throughout the process \cite{jack2008alzheimer}. The dataset consists of a total of 11651 visits over 1346 patients with 6 months intervals. For each patient, 21 variables are collected and processed, including 16 time-varying features (brain function, cognitive tests) and 5 static features (background, demographics). 3 diagnose labels are assigned by doctors at each visit for the patient, including control normal (CN), Mild Cognitive Impairment (MCI), and AD, which indicates the severity of how AD symptoms have progressed on each patient.

\subsubsection{PPMI Dataset}

Parkinson's Progression Markers Initiative (PPMI) is a longitudinal study aiming to evaluate patients' progression on Parkinson's disease (PD) based on biomarkers \cite{marek2011parkinson}. The dataset consists of a total of 13685 visits over 2145 patients with irregular time intervals. For each patient, 79 features based on motor and non-motor symptoms are collected, including cognitive assessment, lab tests, demographic information, and biospecimens. Since the dataset does not provide a diagnosis label per visit for each patient, we use Hoehn and Yahr (HY) scores as labels for our evaluation. HY scores, ranges from 0 to 5, indicate the severity of patients' symptoms of Parkinson's disease. We use the mean and last occurrence carried forward method to impute missing values.

\subsection{Baselines}

We compare our proposed model to several state-of-the-art methods, ranged from vanilla RNNs to multi-layer attention models. Since here we consider disease progression modeling under both supervised and unsupervised settings, we adjusted the architecture of the baseline models to fit the objective accordingly. For baselines that cannot be modified interchangeably, we did not collect the result under the corresponding setting. For all experiments, we use k-means clustering on the hidden representations from the last layer to report the clustering performance. 

\begin{itemize}
  \item \textbf{RNN} \cite{mikolov2010recurrent}: A single RNN cell with an additional layer of feed-forward neural network. The model is trained with cross-entropy loss and reconstruction objective accordingly. 
  \item \textbf{Bi-LSTM} \cite{huang2015bidirectional}: Similar to RNN model, a Bi-directional LSTM is used with a reconstruction objective, the model takes both directions of the sequence data into account and is showed to capture richer information compare to single direction.
  \item \textbf{RETAIN} \cite{choi2016retain}: An interpretable deep learning model that is based on recurrent neural network and reverse time attention mechanism. The RETAIN model learns the importance of hospital records through attention weights. We modify the last layer of RETAIN and train the model based on the prediction and reconstruction objective.
   \item \textbf{Dipole} \cite{ma2017dipole}: A interpretable bidirectional recurrent neural network that employs attention mechanism to leverage both past and future visits. We use concatenation-based attention mechanism for testing and, similar to RETAIN, we adjust the last layer of the model accordingly.
  \item \textbf{StageNet} \cite{gao2020stagenet}: A recent risk prediction model that learned to extract disease progression patterns during training and leveraged modified LSTM cell with an attention mechanism. The progression pattern at each timestamp is re-calibrated accordingly using a convolution network.
  \item \textbf{AC-TPC} \cite{lee2020temporal}: A recent deep predictive clustering network that consists of an encoder network, selector, and a predictor. The model is first initialized using a prediction objective and then optimized to train a cluster embedding using the actor-critic algorithm. This method cannot be trained without label information.
  \item \textbf{VAE} \cite{kusner2017grammar}: A vanilla variational autoencoder model using a LSTM cell as encoder and trained with prediction and variation objective respectively. Note that this baseline method can be served as an ablation example against our proposed method. 
  \item \textbf{Memory Network}: A vanilla global-level memory network with reading and writing mechanism described previously. The network reads and writes the EHR sequence directly and the k-means algorithm is applied directly to the hidden memory representation.
  \item \textbf{$\text{TC-EMNet}^{-u}$}: Unsupervised version of TC-EMNet. When the training label is not available, only a global-level memory network is used to produce memory representation. We also train the model for the prediction task and set it as an ablation example against supervised version of TC-EMNet.
  \item \textbf{$\text{TC-EMNet}^{-s}$}: Supervised version of TC-EMNet. When the training label is available, a patient-level memory network is used to combine with a global-level memory network to produce target-aware memory representations. 
\end{itemize}

\begin{table}[h]
  \centering
  \caption{Hyperparameter Searching Space}
  \label{hyper}
  \begin{tabular}{cc}
    \hline
    Hyperparameter & Range\\
    \hline
    hidden size & $[32,64,128,256]$\\
    latent variable size & $[32,64,128,256]$\\
    x & $[500, 700, 900]$ \\
    learning rate & $[1e-2,1e-3,1e-4,1e-5]$\\
    batch size & $[64, 128]$\\
    \hline

\end{tabular}
\end{table}
\begin{figure*}
\centering
\subcaptionbox{Bilstm}{\includegraphics[width=0.24\textwidth]{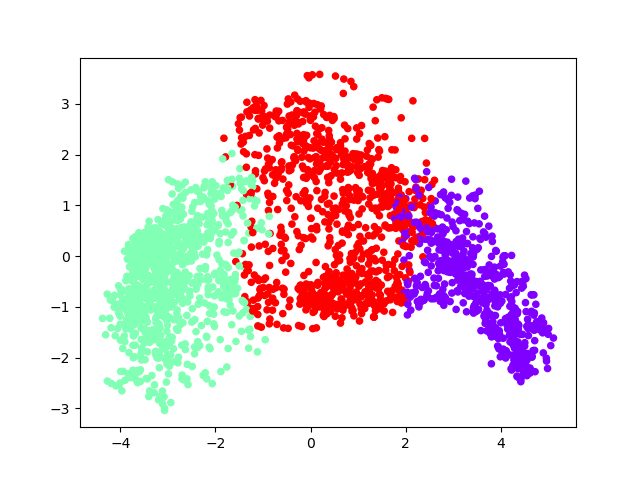}}
\hfill 
\subcaptionbox{StageNet}{\includegraphics[width=0.24\textwidth]{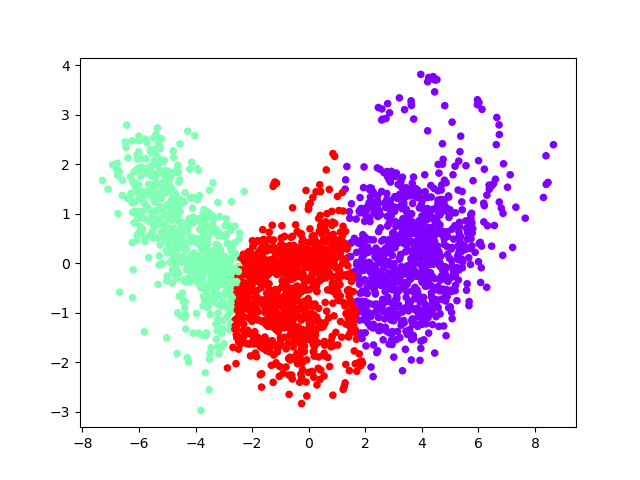}}%
\hfill 
\subcaptionbox{Dipole}{\includegraphics[width=0.24\textwidth]{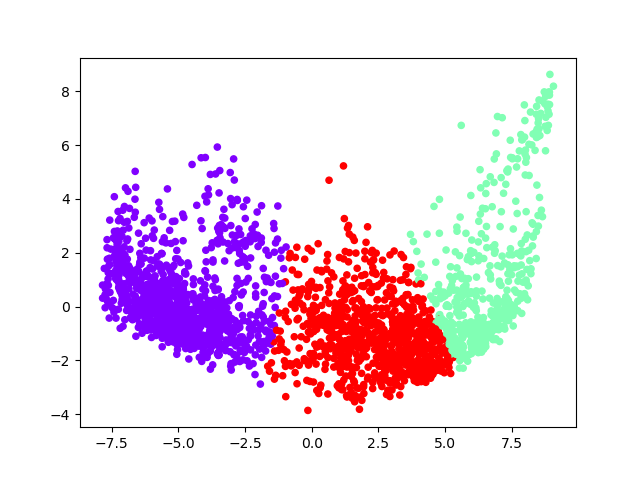}}%
\hfill
\subcaptionbox{Ours}{\includegraphics[width=0.24\textwidth]{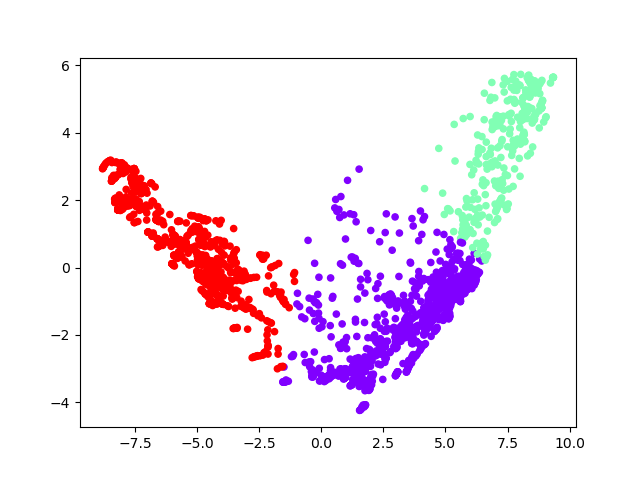}}%
\hfill 
\subcaptionbox{Bilstm}{\includegraphics[width=0.24\textwidth]{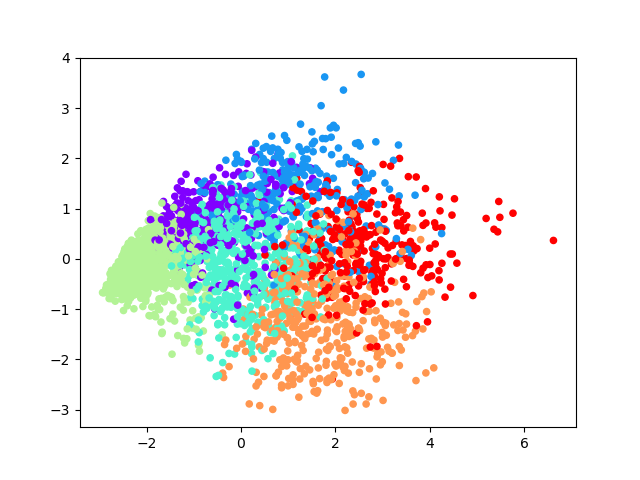}}
\hfill 
\subcaptionbox{StageNet}{\includegraphics[width=0.24\textwidth]{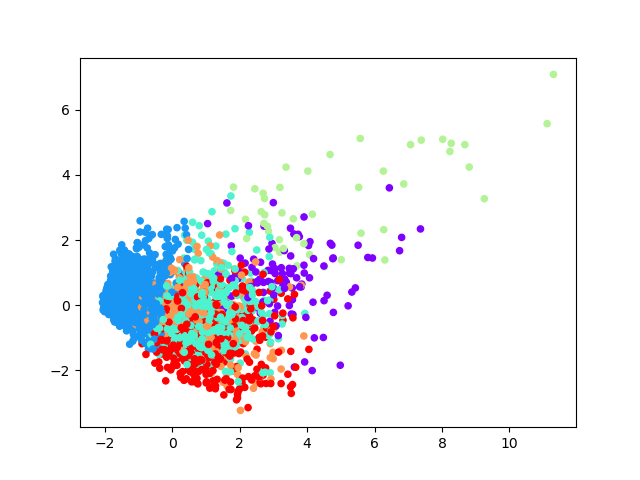}}%
\hfill 
\subcaptionbox{Dipole}{\includegraphics[width=0.24\textwidth]{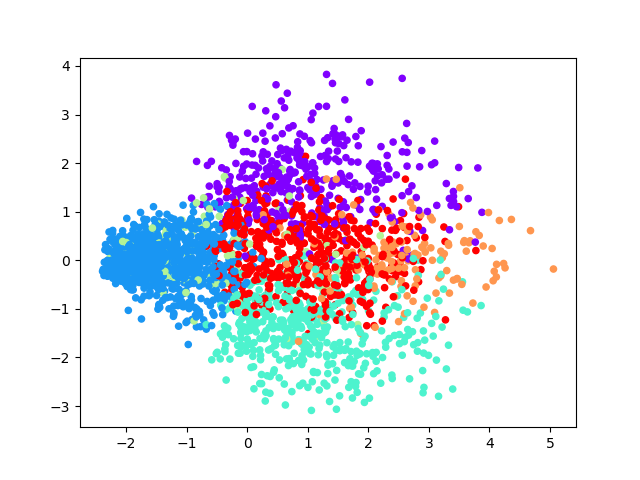}}%
\hfill
\subcaptionbox{Ours}{\includegraphics[width=0.24\textwidth]{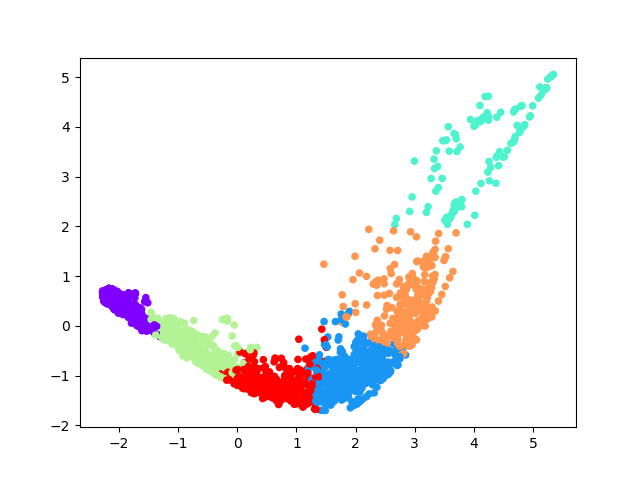}}%
\hfill 
\caption{Visualization of the clusters for ADNI (first row) and PPMI (second row) using PCA: Bilstm (1st column), StageNet (2nd column), 
Dipole (3rd column), Ours (4th) column). }
\label{result_compare}
\end{figure*}

\begin{table*}[t]
    \centering
     \caption{Most significant features in each cluster measured by first order gradient for ADNI and PPMI dataset.}
    \begin{tabular}{ccccccccc}

    \multicolumn{9}{c}{ADNI Dataset}\\
        \hline
        & \multicolumn{8}{c}{Features}\\
        \hline
       Cluster \RomanNumeralCaps {1}& \multicolumn{2}{c}{RAVLT\_learning} & \multicolumn{2}{c}{Ventricles} & \multicolumn{2}{c}{WholeBrain} & \multicolumn{2}{c}{ICV}  \\ 
              & \multicolumn{2}{c}{RAVLT\_perc\_forgetting}   & \multicolumn{2}{c}{RAVLT\_forgetting} & \multicolumn{2}{c}{ADAS13} & \multicolumn{2}{c}{RAVLT\_immediate} \\ 
        \hline
        Cluster \RomanNumeralCaps{2} & \multicolumn{2}{c}{ICV} & \multicolumn{2}{c}{RAVLT\_perc\_forgetting} & \multicolumn{2}{c}{ADAS13} & \multicolumn{2}{c}{Ventricles} \\ 
                 & \multicolumn{2}{c}{serial} & \multicolumn{2}{c}{RAVLT\_immediate} & \multicolumn{2}{c}{CDRSB} \\ 
        \hline
        Cluster \RomanNumeralCaps{3} & \multicolumn{2}{c}{RAVLT\_perc\_forgetting} & \multicolumn{2}{c}{serial

} & \multicolumn{2}{c}{ICV}   & \multicolumn{2}{c}{RAVLT\_learning}\\ 
                & \multicolumn{2}{c}{Entorhinal} & \multicolumn{2}{c}{Hippocampus} & \multicolumn{2}{c}{Ventricles} & \multicolumn{2}{c}{WholeBrain}\\ 
        \hline
        \multicolumn{9}{c}{}\\
      
        \multicolumn{9}{c}{PPMI Dataset}\\
          \hline
        & \multicolumn{8}{c}{Features}\\
        \hline
       Cluster \RomanNumeralCaps {1}& \multicolumn{2}{c}{Global Spontaneity of Movement} & \multicolumn{2}{c}{Speech} & \multicolumn{2}{c}{Anxious Mood} & \multicolumn{2}{c}{Arising from Chair}  \\ 
              & \multicolumn{2}{c}{Right leg}   & \multicolumn{2}{c}{Getting Out of Bed} &  \multicolumn{2}{c}{Pronation-Supination (left)} & \multicolumn{2}{c}{} \\ 
        \hline
        Cluster \RomanNumeralCaps{2} & \multicolumn{2}{c}{Posture} & \multicolumn{2}{c}{Rest tremor amplitude} & \multicolumn{2}{c}{Dopamine} & \multicolumn{2}{c}{Rigidity} \\ 
                 & \multicolumn{2}{c}{Saliva + Drooling} & \multicolumn{2}{c}{Anxious Mood} & \multicolumn{2}{c}{Global Spontaneity of Movement} \\ 
        \hline
        Cluster \RomanNumeralCaps{3} & \multicolumn{2}{c}{Postural Stability} & \multicolumn{2}{c}{Cognitive Impairment
} & \multicolumn{2}{c}{Rest Tremor Amplitude}   & \multicolumn{2}{c}{Pronation-Supination (left)}\\ 
                & \multicolumn{2}{c}{Dopamine} & \multicolumn{2}{c}{Standing} & \multicolumn{2}{c}{Rigidity} & \multicolumn{2}{c}{}\\ 
        \hline
      
       Cluster \RomanNumeralCaps {4}& \multicolumn{2}{c}{Pronation-Supination (left)} & \multicolumn{2}{c}{Standing} & \multicolumn{2}{c}{Postural Stability} & \multicolumn{2}{c}{Chewing}  \\ 
              & \multicolumn{2}{c}{Cognitive Impairment}   & \multicolumn{2}{c}{Dopamine} & \multicolumn{2}{c}{Right Hand} & \multicolumn{2}{c}{} \\ 
        \hline
        Cluster \RomanNumeralCaps{5} & \multicolumn{2}{c}{Dopamine} & \multicolumn{2}{c}{Cognitive Impairment} & \multicolumn{2}{c}{ Hallucinations} & \multicolumn{2}{c}{Chewing} \\ 
                 & \multicolumn{2}{c}{Dressing} & \multicolumn{2}{c}{Pronation-Supination (left)} & \multicolumn{2}{c}{Arising from Chair} \\ 
        \hline
        Cluster \RomanNumeralCaps{6} & \multicolumn{2}{c}{Rigidity} & \multicolumn{2}{c}{Serial

} & \multicolumn{2}{c}{Rigidity}   & \multicolumn{2}{c}{Standing}\\ 
                & \multicolumn{2}{c}{Apathy} & \multicolumn{2}{c}{Constipation Problems} & \multicolumn{2}{c}{Cognitive Impairment} & \multicolumn{2}{c}{Dopamine}\\ 
        \hline
    \end{tabular}
    \label{features}
\end{table*}

\begin{table}[h]
    \centering
     \caption{Complexity comparison between models}
    \begin{tabular}{ cc } 
    \hline
    Model & \# of trainable parameters \\
    \hline
    Dipole & 279k \\ 
    StageNet & 283k \\ 
    AC-TPC & 143k \\ 
    $\text{TC-EMNet}^{-u}$ & 163k \\ 
    $\text{TC-EMNet}^{-s}$ & 174k \\ 
    \hline
    \end{tabular}
    \label{complexity}
\end{table}
\subsection{Model Training and Implementation Details}

As mentioned previously, our proposed network is continuous and differentiable. We can train the network using stochastic optimization techniques. All neural networks in the proposed network are feed-forward networks. We implemented our solution using Pytorch \cite{paszke2019pytorch} and trained the model on a single Nvidia Volta V100 GPU with 16GB memory. We adopt gradient accumulation when dealing with out-of-memory problems. We select hyperparameters through random search as shown in table \ref{hyper}. 
For our model, we set both hidden size and latent variable size to be 128. We adopt Adam optimizer with a learning rate of $1e-3$. The model is trained with batch size $32$ for $70$ epochs. $x$ is set to $700$. We split the dataset into training, validation, and testing set with a ratio of $3/1/1$ and report the performance of $5$ fold cross-validation for both datasets.
A detailed description of the optimization process of our proposed framework can be found in Algorithm \ref{algo}. The average running time of our proposed framework on both datasets is about 2 hours for cross-validation.
For the implementation of other baseline methods, we implement RNN and Bi-lstm methods with Pytorch. We adopt implementations from Pyhealth \cite{zhao2021pyhealth} for RETAIN, Dipole, and StageNet. And we adopt implementation from \cite{lee2020temporal} for AC-TPC. All baseline methods share the same hyperparameter searching space.

\subsection{Evaluation Metrics}

To evaluate the clustering performance of our model, we use purity score (purity), normalized mutual information (NMI) \cite{vinh2010information}, and adjusted rand index (ARI) \cite{hubert1985comparing}. Purity score is ranged between $0$ to $1$, indicating the extent to which a cluster is consist of single class. NMI ($0$ to $1$) represents the mutual information between each clusters with $1$ being perfect clustering. ARI derives from the Rand index and measures the percentage of the correct cluster assignment. Mathematically, the metrics can be expressed as follows:
\begin{equation}
    \begin{aligned}
      &  \textbf{Purity} = \frac{1}{N}\sum^{j}\max|c_{j} \cap l_{j}|, \\
      &  \textbf{NMI} = \frac{2\cdot\textbf{I}(c_{j}, l_{j})}{[H(C)+H(L)]},\\
      & \textbf{ARI} = \frac{RI - E(RI)}{\max({RI}) - E(RI)},
    \end{aligned}
\end{equation}

where $N$ is the total number of samples, $c_{j}$ and $l_{j}$ denotes the cluster assignment and true label respectively, $I(\cdot)$ is the mutual information function and $H(\cdot)$ is the entropy, $E(\cdot)$ and \textit{RI} are the expectation value and Rand index accordingly.

\section{Results}

\subsection{Clustering Performance}

A quantitative comparison of the clustering performance on ADNI and PPMI dataset is shown in table \ref{ADNI_results} and table \ref{PPMI_results} respectively. We set the cluster assignments to the number of class/diagnosis for each dataset, i.e. $3$ for ADNI (diagnosis label) and $6$ (NHY score) for PPMI. We want the model to identify the individual disease stages both when there is only limited knowledge known to a certain disease, i.e class/diagnosis is not available and when diagnosis label is available, and thus provide insightful and interpretable information to help discover corresponding treatment to individual treatment. We compare our proposed method with the aforementioned baselines in terms of clustering performance. It is clear that our method has demonstrated competitive performance against all baseline methods across all evaluation metrics for both datasets. We note that it is generally difficult to identify clusters without the presence of label information as indicated by low NMI and RI scores. However, TC-EMNet outperforms baseline by a large margin in terms of NMI and RI scores when clustering with label. Training under supervised setting yields significantly better clustering performance compared to training under unsupervised setting. This is due to fact that the correlation between diagnosis and input features is encoded into each hidden representation. Although AC-TPC has better performance in terms of RI on the PPMI dataset. The method relies on pre-training the model with over $1000$ epochs, which could result in the model memorizing the input data. Both Dipole and StageNet have comparable performance. However, it is worth mentioning that both models have leveraged attention over multi-layer RNNs, which introduces additional complexity to the model. A detailed comparison between the trainable parameters is shown in table~\ref{complexity}. Furthermore, we find that when training with label information, RI score can be negatively impacted compared to training without labels. Such phenomena are observed for multiple baseline methods. One explanation could be directly leveraging label information overwhelms the training process since labels possess strong prediction power compared to input features, making the model more biased towards dominated class when dealing with imbalanced datasets; thus, RI may drop as there are more false positives and false negatives. It also can be observed that leveraging external memory effectively captures long-term information and the TC-EMNet is capable of learning complexity from the input data. The patient-level memory network constructively binds with the global-level memory network to produce more comprehensive memory representations.

\begin{figure}
  \centering
  \includegraphics[width=\linewidth]{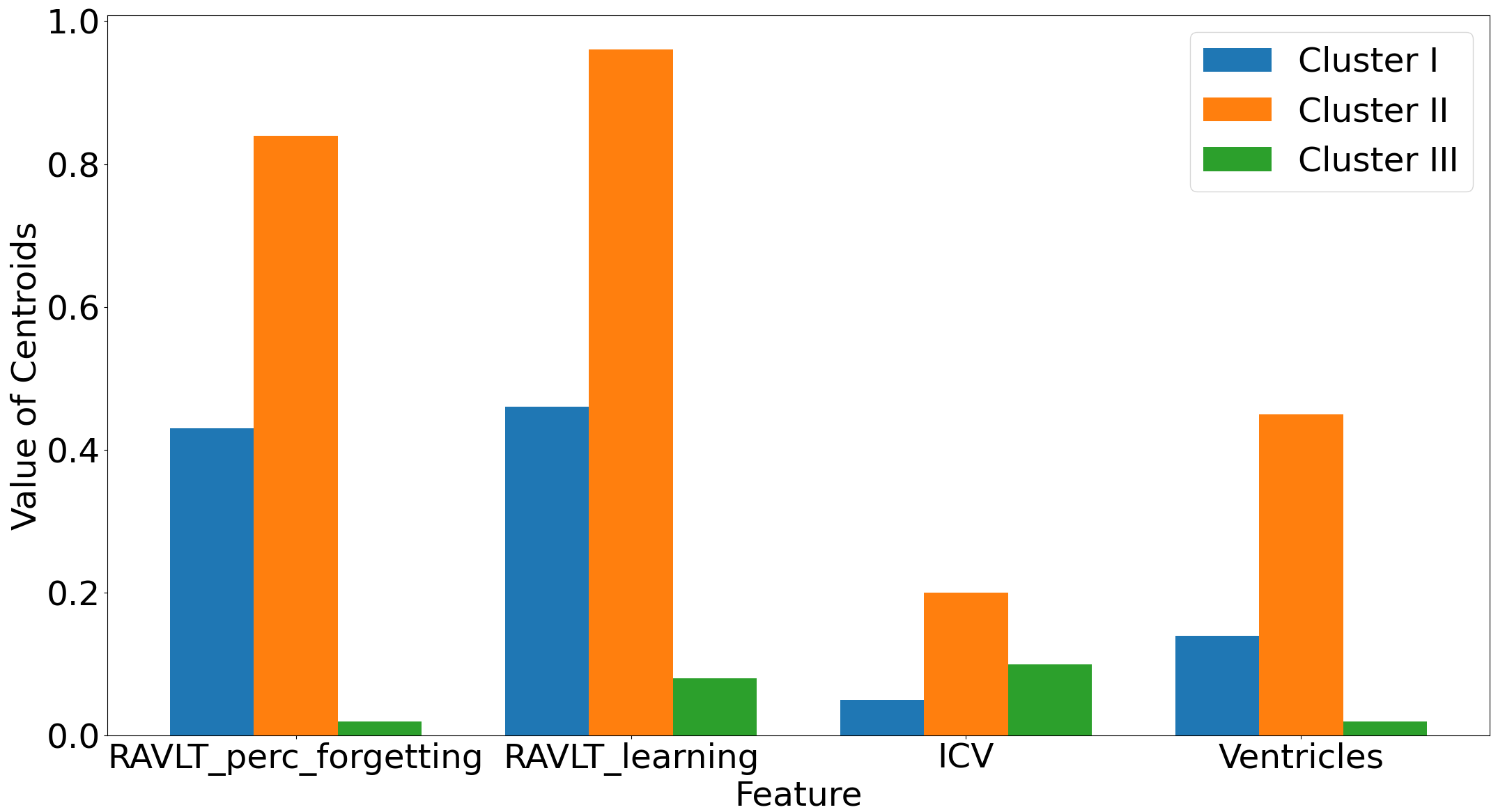}
  \caption{
        Significant feature values of cluster centroids on ADNI dataset. The distribution of clusters is very different, which means distinct subtypes.
  }
  \label{feat1}
\end{figure}

\begin{figure}
  \centering
  \includegraphics[width=\linewidth]{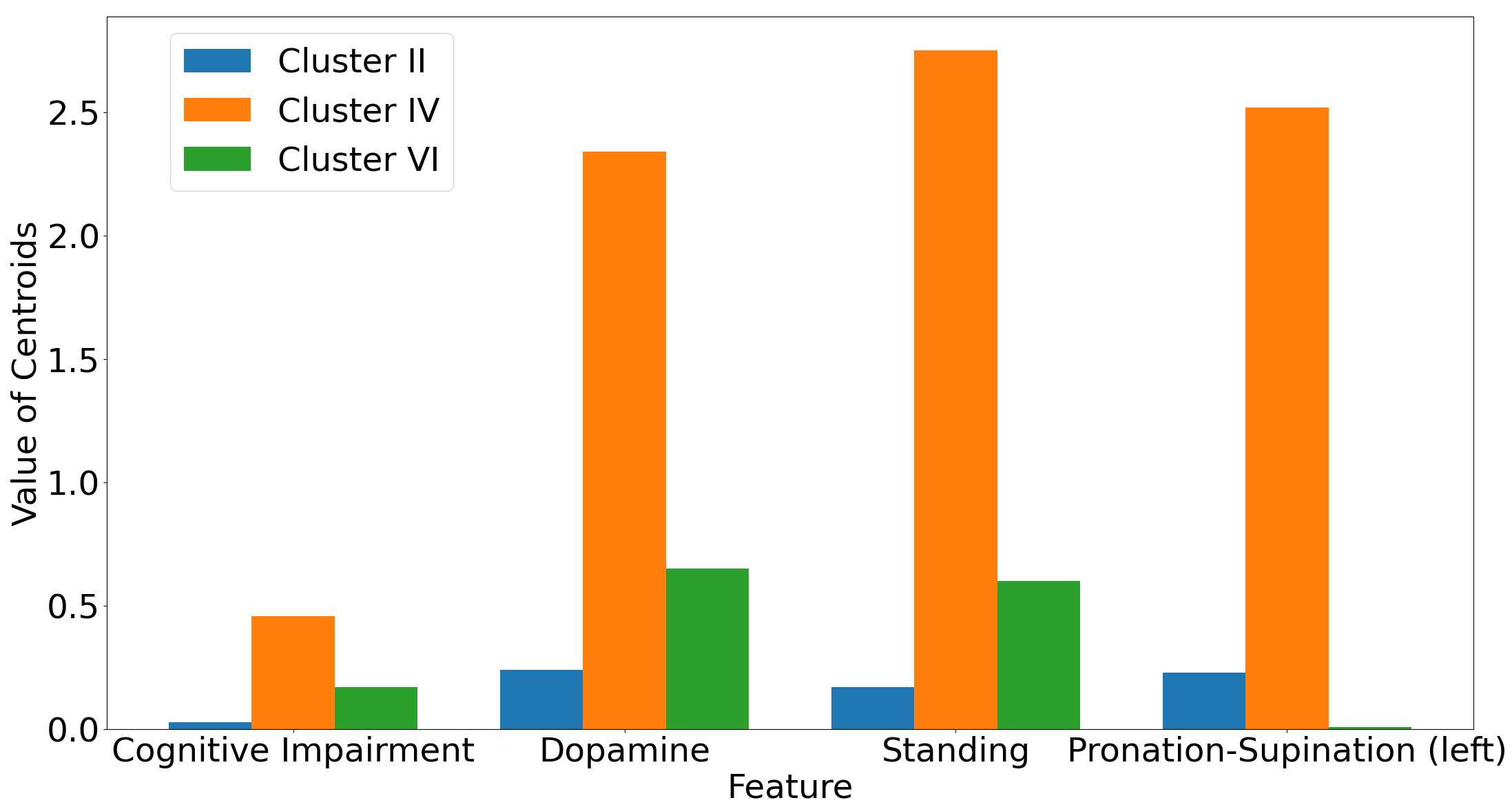}
  \caption{
        Significant feature values of cluster centroids on PPMI dataset. The distribution of clusters is very different, which means distinct subtypes.
  }
  \label{feat2}
\end{figure}

\subsection{Disease Stage}

In order to interpret the disease stages and progression patterns found by TC-EMNet. We first selected three baseline models that have comparable performance against TC-EMNet and visualized the hidden representations in $2D$ space using PCA \cite{martinez2001pca}. The results are shown in Fig \ref{result_compare}. We observed that in general most methods can produce distinct clusters for the ADNI dataset. However, for PPMI dataset, most baseline methods failed at producing effective clusters, whereas TC-EMNet produces distinct clustering results. This shows that TC-EMNet is able to constructively model long-term information between each visit in order to find effective representations. Next, we compute feature importance for every cluster based on the weights from the last layer of the network. The results are shown in table \ref{features}. It can be observed that for both datasets, each cluster is determined by a diverse range of features, which means it is easier to identify each patients' progression patterns through observation. We also compute the centroid values for each cluster and plot the distribution in Fig \ref{feat1}, \ref{feat2} for ADNI and PPMI datasets respectively. For ADNI dataset, our proposed model has determined significant features such as RAVLT\_learning, RAVLT\_perc\_forgetting, ICV, ventricles. Rey's Auditory Verbal Learning Test (RAVLT) scores are helpful in testing episodic memories and are very important indicators in identifying a patient's progression in Alzheimer's disease \cite{moradi2017rey}. In particular, the learning test (RAVLT\_learning) and percent forgetting test (RAVLT\_perc\_forgetting) are highly correlated and thus become crucial biomarkers for early detection in AD. It can be observed in Fig \ref{feat1} that three clusters produced by our model have wide distribution for RAVLT testing values, which suggests three different patient subtypes. As for PPMI dataset, our model has found that the dopamine dysregulation syndrome (Dopamine) is a significant feature in identifying clusters. Studies have discovered that under clinical settings early characterization of Dopamine can aid the treatment for motor and non-motor complications for Parkinson's disease \cite{evans2004dopamine}. There are also studies that showed that cognitive impairment (Cognitive impairment) is a strong indicator for Parkinson's disease. Difference in cognitive impairment scores can reflect advanced progression in PD \cite{verbaan2007cognitive}.

\section{Conclusion}

In this paper, we propose TC-EMNet for disease progression modeling on time-series data. TC-EMNet leverages VAE to model data irregularity and an external memory network to capture long-term dependency. We developed TC-EMNet to perform patient clustering/subtyping under both supervised and unsupervised settings. Under supervised setting, TC-EMNet leverages a dual memory network architecture to extract target-aware information from diagnosis to compute patient representations. Throughout the experiment on two real-world datasets, we showed that our model outperforms state-of-the-art methods and is able to identify interpretable disease stages that are clinically meaningful. TC-EMNet yields competitive clustering performance with limited complexity. In the real-world clinical setting, we hope that our model could help physicians identify patients' progression patterns and discover potential disease stages to gain more understanding about chronic and other heterogeneous diseases.

\section*{Acknowledgment}
This paper was funded in part by the National Science Foundation under award number CBET-2037398.

\bibliographystyle{IEEEtran}
\bibliography{reference}

\end{document}